\documentclass[conference]{IEEEtran}
\IEEEoverridecommandlockouts

\usepackage{amsmath,amssymb}

\usepackage{times}
\usepackage{bm}
\usepackage{algorithm}
\usepackage{xargs}[2008/03/08]
\usepackage{amssymb}
\usepackage{mathrsfs}
\usepackage{graphicx}
\usepackage{rotating,subfigure}
\usepackage[flushleft]{threeparttable}
\usepackage{multirow}
\usepackage{enumitem} 
\usepackage[normalem]{ulem}
\useunder{\uline}{\ul}{}
\usepackage{multirow}

\usepackage{star}

\usepackage{cite}
\usepackage{amsmath,amssymb,amsfonts}
\usepackage{algorithmic}
\usepackage{graphicx}
\usepackage{textcomp}
\usepackage{xcolor}
\def\BibTeX{{\rm B\kern-.05em{\sc i\kern-.025em b}\kern-.08em
		T\kern-.1667em\lower.7ex\hbox{E}\kern-.125emX}}

\begin{document}

\title{M2A: Motion Aware Attention for Accurate Video Action Recognition }

\author{\IEEEauthorblockN{Brennan Gebotys}
	\IEEEauthorblockA{\textit{Systems Design Engineering} \\
		\textit{University of Waterloo}\\
		bmgebotys@uwaterloo.ca}
	\and
	\IEEEauthorblockN{Alexander Wong}
	\IEEEauthorblockA{\textit{Systems Design Engineering} \\
		\textit{University of Waterloo}\\
		a28wong@uwaterloo.ca}
	
	\and
	\IEEEauthorblockN{David A. Clausi}
	\IEEEauthorblockA{\textit{Systems Design Engineering} \\
		\textit{University of Waterloo}\\
		dclausi@uwaterloo.ca}
}

\maketitle

\begin{abstract}
	Advancements in attention mechanisms have led to significant performance improvements in a variety of areas in machine learning due to its ability to enable the dynamic modeling of temporal sequences. A particular area in computer vision that is likely to benefit greatly from the incorporation of attention mechanisms in video action recognition. However, much of the current research's focus on attention mechanisms have been on spatial and temporal attention, which are unable to take advantage of the inherent motion found in videos. Motivated by this, we develop a new attention mechanism called Motion Aware Attention (M2A) that explicitly incorporates motion characteristics. More specifically, M2A extracts motion information between consecutive frames and utilizes attention to focus on the motion patterns found across frames to accurately recognize actions in videos. The proposed M2A mechanism is simple to implement and can be easily incorporated into any neural network backbone architecture. We show that incorporating motion mechanisms with attention mechanisms using the proposed M2A mechanism can lead to a +15\% to +26\% improvement in top-1 accuracy across different backbone architectures, with only a small increase in computational complexity. We further compared the performance of M2A with other state-of-the-art motion and attention mechanisms on the Something-Something V1 video action recognition benchmark.  Experimental results showed that M2A can lead to further improvements when combined with other temporal mechanisms and that it outperforms other motion-only or attention-only mechanisms by as much as +60\% in top-1 accuracy for specific classes in the benchmark. We make our code available at: \url{https://github.com/gebob19/M2A}.
\end{abstract}


\section{Introduction}

Attention is a technique that mimics cognitive attention by either enhancing or diminishing parts of the input data to achieve improved performance. Attention in the context of neural networks has led to improved results across a variety of fields including natural language processing \cite{Vaswani2017}, image classification \cite{Dosovitskiy2020}, image segmentation \cite{Zhong2020}, and others due to its ability to dynamically model sequences. A field that will likely benefit greatly from the application of attention, is video action recognition. Current research has investigated using attention to attend across the frames of the video (temporal attention) and patches of the current image (spatial attention) \cite{Wu2021,He2020,Bertasius2021}. However, these methods attend only to visual features and do not utilize the motion between frames which is an important feature specific to videos. 

\begin{figure}[t]
	\includegraphics[scale=0.43]{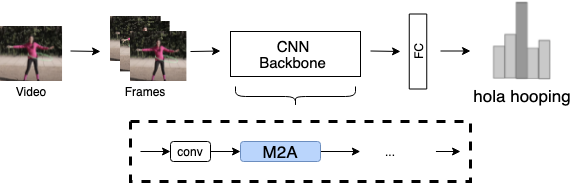}
	\caption{Overview of the proposed M2A mechanism within the context of deep-learning-driven video action recognition.  The M2A mechanism can be added to any deep neural network backbone architecture to enable explicit incorporation of motion characteristics.  Frames are first sampled from the video and are then processed by the network.}
	\label{fig:network}
\end{figure}

Motivated by this, we develop a motion aware attention mechanism (M2A) that leverages both motion information and attention for accurate video action recognition. M2A works by first extracting motion information between consecutive frames and then utilizes attention to focus on relevant motions found across frames. We designed M2A to be easy to implement and can be easily inserted into any backbone architecture. For example, Figure \ref{fig:network} shows how M2A could be inserted into a general convolution neural network (CNN) backbone.

To understand how motion can help improve the performance of attention-based methods we perform ablation studies across three CNN backbones including 2D-ResNet \cite{He2016}, 2D-MobileNet\cite{Sandler2018}, and I3D-inflated-ResNet \cite{Carreira2017}. Furthermore, we use Grad-CAM \cite{Selvaraju2020} to visualize how the model’s focus changes when using motion, attention, and both motion and attention. We also compare the performance of M2A to other state-of-the-art (SOTA) mechanisms on the video action recognition benchmark SomethingSomething V1 \cite{Goyal2017}. Finally, to understand if further performance improvements can be achieved, we investigate incorporating M2A with other temporal mechanisms and the change in performance when state-of-the-art motion-only and attention-only mechanisms are combined. 

\section{Previous Work}

\vspace{2mm}
\textbf{Attention Mechanisms.} 
Since attention showed impressive results in natural language processing tasks \cite{Vaswani2017}, using attention for video action recognition has become popular. Typically, these attention mechanisms are used to incorporate temporal information in 2D backbone networks (e.g, ResNets and MobileNets) which otherwise process the frames independently. One of the first works to use attention for video action recognition was Non-Local Networks \cite{Wang2018} which proposed using a self-attention block across the frames to improve temporal modeling. TAM \cite{Wu2021} further extended this idea by using dynamic convolutions and attention. Recently, TimeSformer \cite{Bertasius2021} processed the frames with a transformer network using a patch-based approach. Though these methods have shown impressive results, they do not incorporate any motion information, which is a key feature of videos. 

\vspace{2mm}
\textbf{Motion Mechanisms.}  
Previous research has shown that a valuable feature for achieving high accuracy for video recognition is the relative motion across time (e.g, optical flow) \cite{Simonyan2014}. Furthermore, experiments that use motion information and visual features as input have been shown to outperform visual features alone \cite{Lin2020, Carreira2017}. Due to this, there has been much research on how to design motion mechanisms for video action recognition. TEA \cite{Li2020} investigated different architectures and mechanisms which computed the difference between consecutive frames. TEIN \cite{Liu2019} investigated scaling the input by the motion information. TDN \cite{tdn} investigated multiple motion mechanisms and utilized pooling operations to leverage motion information at multiple spatial scales. However, these methods have not investigated how to utilize attention mechanisms for further performance improvements.  

\begin{figure}[t]
	\centering
	\includegraphics[scale=0.6]{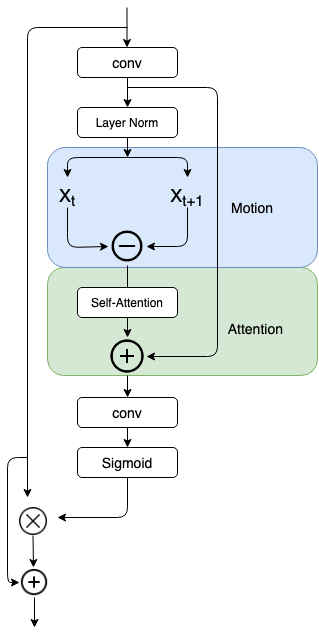}
	\caption{The proposed M2A mechanism consists of a motion block (shown in blue) which extracts motion information across consecutive frames and an attention block (shown in green) which focuses on relevant motion patterns found across frames.}
	\label{fig:block}
\end{figure}

\section{Methodology}

In this section, we introduce the methodology of our motion aware attention (M2A) mechanism. We define the input activation as $X \in \mathbb{R}^{T \times H \times W \times C}$ where $T$ is the temporal dimension and $H$, $W$, and $C$ are the height, width, and the number of channels.


In the first stage of M2A, following the standard practice of video action recognition mechanisms, we apply a convolution operation to our input $X$ to reduce the number of channels by a factor of $R$. This allows us to compute future operations efficiently (we use $R=8$ throughout the paper). This produces a new vector, $X_t \in \mathbb{R}^{T \times H \times W \times C/R}$. Following standard attention-based practice, we then apply a layer normalization operation. 

\textbf{Motion Block:} We then compute a shifted representation of $X_t$, which we denote as $X_{t+1}$, by shifting the temporal axis of $X_{t}$ to the left and filling the last index with the values of the first frame. Next, to extract motion information we compute the difference between $X_{t+1}$ and $X_{t}$. Specifically, this represents the difference in activation values between consecutive frames, emulating the motion between frames.

\textbf{Attention Block:} Next, to help focus on motion patterns found across frames, we flatten the frames to create a vector with shape ${T \times (H * W * C/R)}$ and apply self-attention across the time axis. This is followed by a skip connection of the convolved input. 

Finally, based on previous research \cite{tdn, Li2020, Liu2019}, we excite and inhibit relevant and non-relevant features in the original input by applying a convolution operation to increase the number of channels from $C/R$ back to $C$, followed by the $sigmoid$ activation function, and an element-wise multiplication on the original input, followed by a  skip connection. 

\section{Experiments}

\begin{figure*}[t]
	\centering
	\includegraphics[width=\textwidth]{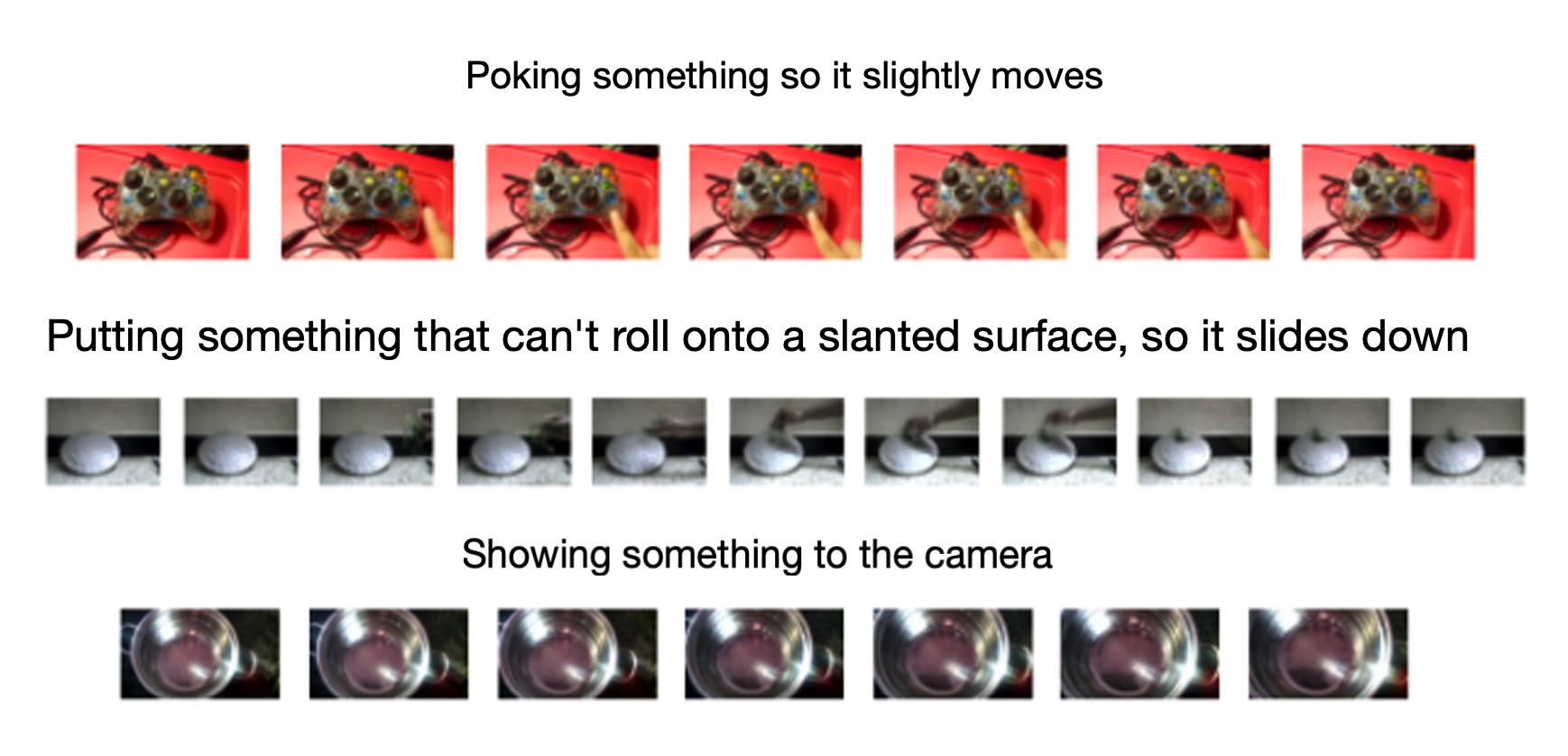}
	\caption{Example video sequences from the Something-Something V1 dataset and their corresponding labels.}
	\label{fig:ss_examples}
\end{figure*}

In this section, we report the results of our experiments. Following standard practice, the temporal mechanism is inserted after the first convolution of each ResNet/MobileNet block. Furthermore, all backbones are initialized with ImageNet \cite{imagenet} pre-trained weights. We performed all our experiments on the Something-Something V1 (SSv1) \cite{Goyal2017} dataset, a standard video action recognition benchmark. Example video sequences from the benchmark dataset and their corresponding labels are shown in Fig \ref{fig:ss_examples}. Throughout the experiments, we uniformly sampled 8 frames from the video and use them as input to the model and use a 2D-ResNet18 backbone unless stated otherwise. We also report the number of Giga Multiply-Accumulate operations (GMACs) per video (GMACs/video) as a measure of model efficiency, and the top-1 and top-5 accuracy (Top-1 Acc and Top-5 acc) as a measure of the model's performance. 

\subsection{Ablation Experiments}

To understand how motion and attention contribute to the model's performance and if the results generalize across different backbones we run ablation studies across three backbones: 2D-ResNet18 \cite{He2016}, 2D-MobileNetV2 \cite{Sandler2018}, and I3D-inflated-Resnet18 \cite{Carreira2017, chen2020deep}. Specifically, we compared each backbone with no temporal mechanism (None), with M2A but the motion block is removed (M2A-Attention), with M2A but the attention block is removed (M2A-Motion), and with the full M2A mechanism (i.e, with both motion and attention blocks) (M2A). In the tables, the percentage improvement from None is noted in brackets and the best accuracy is bolded.

\begin{table}[t]
	\caption{Ablation study of different temporal mechanisms using a 2D-ResNet18 backbone.}
	\def\arraystretch{1.25}
	\centering
	\begin{tabular}{|l|l|l|l|}
		\hline
		Mechanism           & GMACs/video & Top-1 Acc  & Top-5 Acc  \\ \hline
		None                & 14.57       & 13.9       & 38.9       \\ \hline
		M2A-Attention           & 14.79       & 16.9 (+3\%)  & 42.0 (+3\%)  \\ \hline
		M2A-Motion              & 14.79       & 28.9 (+15\%) & 56.7 (+17\%) \\ \hline
		M2A & 14.81       & \textbf{34.7 (+20\%)} & \textbf{63.4 (+24\%)} \\ \hline
	\end{tabular}
	\label{table:abl-resnet2d}
\end{table}

\begin{table}[t]
	\caption{Ablation study of different temporal mechanisms using a 2D-MobileNetV2 backbone.}
	\def\arraystretch{1.25}
	\centering
	\begin{tabular}{|l|l|l|l|}
		\hline
		Mechanism           & GMACs/video & Top-1 Acc     & Top-5 Acc     \\ \hline
		None                & 2.55        & 13.8          & 37.8          \\ \hline
		M2A-Attention           & 2.58        & 13.3 (-0.5\%) & 37.2 (-0.6\%) \\ \hline
		M2A-Motion              & 2.58        & 32.0 (+18\%)  & 60.6 (+22\%)  \\ \hline
		M2A & 2.58        & \textbf{35.6 (+21\%)}  & \textbf{64.4 (+26\%)}  \\ \hline
	\end{tabular}
	\label{table:abl-mobile}
\end{table}

\subsubsection{2D-ResNet18 Backbone}

We first perform our experiments using the 2D-ResNet18 backbone since it is a popular choice for computer vision tasks and has shown to achieve good results without being very computationally expensive \cite{He2016}. Table \ref{table:abl-resnet2d} shows the results of our ablation study using a 2D-ResNet18 backbone. We see that M2A-Attention improves upon the top-1 accuracy compared to None by +3\%, but it’s unable to achieve a large improvement. This could be because it’s only focusing on similar visual features across frames and is unable to extract motion information. We also see that M2A-Motion outperforms None by +15\%, which shows the importance of extracting motion information for improved video action recognition performance. Lastly, M2A achieves the highest accuracy improvement of +20\%, meaning that using attention to focus on motion patterns across frames is the best compared to using only motion or only attention. More specifically, comparing M2A to M2A-Attention, we see that incorporating motion with attention mechanisms outperforms attention-only mechanisms by +17\% in top-1 accuracy which further supports the idea that incorporating motion mechanisms with attention mechanisms can lead to improved results for video action recognition. We see a similar trend for the top-5 accuracy which shows that these improvements are achieved across multiple metrics. Lastly, we see that including the M2A mechanism only increases the GMACs/video by only +1.6\% showing that M2A is very computationally efficient.

\subsubsection{2D-MobileNetV2 Backbone}

Relative to 2D-ResNet18 backbones, 2D-MobileNetV2 backbones are typically used in more resource-constrained settings where computational power and latency must be kept to a minimum (e.g, mobile and edge devices). Since this is a common setting for video action recognition we perform the same ablation studies using the 2D-MobileNetV2 backbone. Table \ref{table:abl-mobile} shows the results of the ablation experiments. We see that the ablation study follows a similar trend as in the 2D-ResNet18 ablation study which further supports the idea that further accuracy improvements can be achieved by incorporating motion mechanisms with attention mechanisms for video action recognition. Furthermore, we see that M2A achieves a higher top-1 accuracy with the 2D-MobileNetV2 backbone (i.e, 35.6\%) compared to the 2D-ResNet18 backbone (i.e, 34.7\%) while having more than 5 times lower GMACs/video (i.e, 14.81 GMACs/video with a 2D-ResNet18 backbone and 2.58 GMACs/video with a 2D-MobileNetV2). This shows that M2A generalizes across different architectures and is a viable option in resource-constrained settings.

\begin{table}[t]
	\caption{Ablation study of different temporal mechanisms using a I3D-ResNet18 backbone.}
	\label{table:abl-i3d}
	\def\arraystretch{1.25}
	\centering
	
	\begin{tabular}{|l|l|l|l|}
		\hline
		Mechanism           & GMACs/video & Top-1 Acc     & Top-5 Acc     \\ \hline
		None                & 22.52       & 27.0          & 53.5          \\ \hline
		M2A-Attention           & 22.65       & 26.9 (-0.1\%) & \textbf{54.1 (+0.6\%)} \\ \hline
		M2A-Motion              & 22.65       & 26.3 (-0.7\%) & 53.6 (+0.1\%) \\ \hline
		M2A & 22.67       & \textbf{27.1 (+0.1\%)} & 53.5 (0\%)    \\ \hline
	\end{tabular}
\end{table}

\subsubsection{I3D-ResNet18 Backbone}

I3D-inflated-ResNet18 backbones use 3D convolutions instead of 2D convolutions. While 2D CNNs model the frames individually, these 3D CNNs explicitly model the temporal aspect of videos which has been shown to achieve accurate results without requiring any additional temporal mechanisms. To understand if M2A can improve 3D convolution networks, Table \ref{table:abl-i3d} shows the results of the same ablation study but uses an I3D-inflated-ResNet18 backbone. We see only a small improvement using M2A compared to None. This means that M2A is unlikely to further improve 3D CNNs since temporal information is already modelled accurately. However, we also see that using 3D CNNs is much more computationally expensive requiring 22 GMACs/video. Furthermore, if we compare these results to our 2D-MobileNetV2 experiments in Table \ref{table:abl-mobile} we see that M2A achieved a top-1 accuracy of 35.6\% with only 2.58 GMACs/video while using 3D CNNs achieved only 27.0\% top-1 accuracy with approximately 10 times the computational cost. This shows that our M2A mechanism can outperform 3D CNNs while being significantly more computationally efficient. 

\begin{figure*}[t]
	\centering
	\includegraphics[width=\textwidth ]{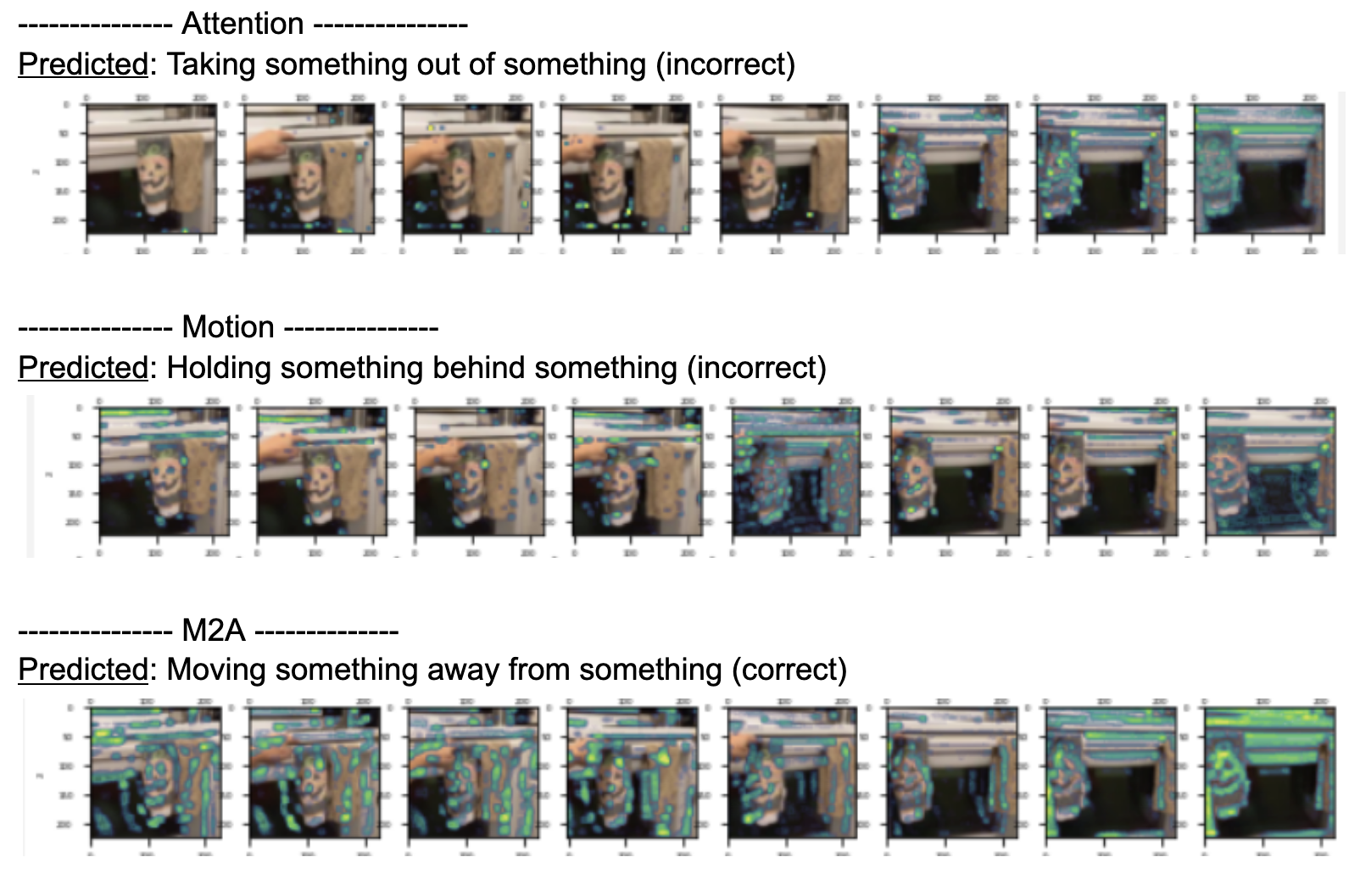}
	\caption{Grad-CAM heatmaps of a ResNet18 backbone on an example video sequence from the Something-Something V1 validation dataset with just the attention mechanism (i.e., M2A-Attention), just the motion mechanism (i.e., M2A-Motion), and the proposed M2A mechanism that incorporates both motion and attention. Images are of the last five frames of an eight-frame video sequence.  We see that when we combine both motion and attention within the proposed M2A mechanism the model pays attention to a large amount of the frames and correctly classifies the video as moving something away from something, whereas it’s unable to when we use just motion or just attention.}
	\label{fig:gradcam}
\end{figure*}

\subsection{Visualizing Model Focus with Grad-CAM \cite{Selvaraju2020} }

To further understand the difference between motion, attention, and both motion and attention, we visualized where the model is focusing using Grad-CAM \cite{Selvaraju2020} heatmaps. Figure \ref{fig:gradcam} shows the resulting heatmaps extracted from the second layer of a 2D-ResNet18 backbone. The blue/green values show where each model focuses. We also include the predicted and ground truth class. 

In the first section of Figure \ref{fig:gradcam} we see the results of using just the attention component of the proposed M2A mechanism (we will refer to that as M2A-Attention). It can be observed that when we only use an attention mechanism to model temporal information, the model only seems to focus on the last few frames. Furthermore, it incorrectly classifies the action as taking something out of something. The second section shows the results of using just the motion component of the proposed M2A mechanism (we will refer to that as M2A-Motion). Here the model doesn't focus on anything specifically and incorrectly classifies it as holding something behind something. In the last section, we see the results from the proposed M2A. We see that when we combine both motion and attention within the proposed M2A mechanism the model pays attention to a large amount of the frames and correctly classifies the video as moving something away from something, whereas it’s unable to when we use just motion or just attention. This shows that combining motion with attention can lead to the model focusing on more important details in the video to correctly classify the action.

\begin{table}[t]
	\caption{Comparison of different state-of-the-art attention mechanisms used in M2A with a 2D-ResNet18 backbone.}
	\label{table:state-of-the-art-attn}	
	\def\arraystretch{1.25}
	\centering
	\begin{tabular}{|l|l|l|}
		\hline
		Attention mechanism   & \multicolumn{2}{c|}{Top-1 Acc} \\ \hline
		M2A-Attention          & \multicolumn{2}{l|}{\textbf{34.7}}      \\ \hline
		TAM \cite{Wu2021}               & \multicolumn{2}{l|}{31.6}      \\ \hline
		S+T Patch ($s$=4) \cite{Bertasius2021} & \multicolumn{2}{l|}{31.9}      \\ \hline
		S+T Patch ($s$=8) \cite{Bertasius2021} & \multicolumn{2}{l|}{34.2}      \\ \hline
	\end{tabular}
\end{table}

\subsection{Using SOTA Attention}

We also investigated if we could improve the performance of M2A by using SOTA attention blocks. Table \ref{table:state-of-the-art-attn} shows the results of M2A using M2A's attention block (M2A-Attention), space and time based attention block (S+T Patch with a patch of size $s \times s$) from \cite{Bertasius2021}, and TAM attention \cite{Wu2021}. We found that M2A's attention performed the best; and that there wasn't a significant performance difference across the other SOTA attention blocks.

\begin{table}[t]
	\caption{Comparison across different state-of-the-art temporal mechanisms.}
	\label{table:state-of-the-art-mod-compar}
	\centering
	\def\arraystretch{1.25}
	\begin{tabular}{|l|l|l|}
		\hline
		Temporal mechanism & GMACs/Video    & Top-1 Acc        \\ \hline
		M2A + TSM            & 14.81 & \textbf{39.3} \\ \hline
		TSM    \cite{Lin2020}          & 14.57 & 39.0 \\ \hline
		M2A             & 14.81    &  34.7    \\ \hline
		TEA     \cite{Li2020}        & 14.83          & 34.3          \\ \hline
		TDN \cite{tdn}            & 15.13          & 28.6          \\ \hline
		TAM   \cite{Wu2021}          & 14.79          & 21.0          \\ \hline
	\end{tabular}
\end{table}

\begin{figure*}[t]
	\centering
	\includegraphics[width=1\textwidth]{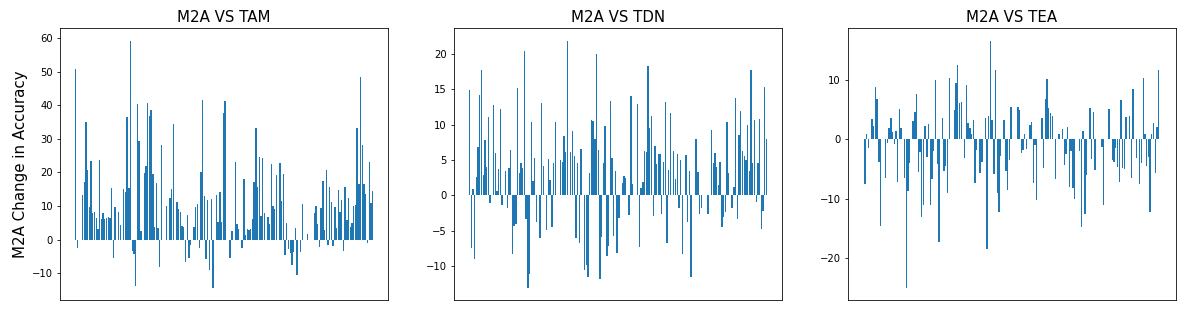}
	\caption{The change in accuracy using M2A compared to other SOTA mechanisms for each class in the SSV1 dataset. Each bar represents a class in the SSV1 dataset (e.g, Showing something to the camera) and the height of the bar represents the difference between M2A and the compared mechanism (e.g, in the M2A VS TAM chart, the largest bar has a height of 60 which means M2A achieves 60\% better accuracy in that class compared to TAM). We see M2A achieves large improvements in most classes when compared to the other SOTA mechanisms.}
	\label{fig:comparison}
\end{figure*}

\subsection{Comparison to SOTA}

Next, we compared M2A's performance to other SOTA temporal mechanisms including, TSM \cite{Lin2020}, TEA \cite{Li2020}, TDN \cite{tdn}, and TAM \cite{Wu2021}. TSM doesn't incorporate any attention or motion information and instead shifts values across consecutive frames, while TEA and TDN are motion-only temporal mechanisms, and TAM is an attention-only temporal mechanism. The TDN and TEA papers involve multiple mechanisms, some of which modify the backbone directly, so to conduct a fair comparison, we only use the individual temporal mechanisms of each method. Specifically, we use the long-term mechanism for TDN and the motion excitation mechanism for TEA. In all our experiments we use the respective paper's publicly available code. Furthermore, to understand if M2A can further improve the performance of other temporal mechanisms, we also compare using both M2A and TSM (M2A + TSM). Table \ref{table:state-of-the-art-mod-compar} shows that M2A can achieve higher accuracy than complex state-of-the-art motion/attention-only mechanisms (TEA, TDN, and TAM). Furthermore, we see that M2A is comparable in terms of GMACs/Video to all the other mechanisms which shows that it is a computationally efficient mechanism. Comparing M2A to TSM, we see that TSM outperforms M2A in top-1 accuracy by approximately 4\%. However, we also see that M2A + TSM outperforms TSM by +0.3\% in top-1 accuracy, showing that M2A is a complementary mechanism that can be combined with other temporal mechanisms to achieve better performance. 

To further understand where M2A outperforms the other SOTA mechanisms, Fig \ref{fig:comparison} shows the difference in top-1 accuracy across all the classes in the SSv1 dataset when M2A is compared to TAM, TDN, and TEA. Specifically, each bar represents a class in the SSv1 dataset (e.g, Showing something to the camera) and the height of the bar represents the difference between M2A and the compared mechanism (e.g, in the M2A VS TAM chart, the largest bar has a height of 60 which means M2A achieves 60\% better accuracy in that class compared to TAM). We see M2A achieves large improvements in most classes when compared to the other SOTA mechanisms, specifically, we see up to +60\%, +20\%, and +10\% top-1 accuracy improvements compared to TAM, TDN, and TEA respectively. Furthermore, we investigated which specific classes had the largest difference. Comparing M2A to TAM, the largest improved classes were:
\begin{itemize}
  \item ``Moving away from something with your camera" (+59.3\%)
  \item ``Approaching something with your camera" (+50.7\%)
\end{itemize}

Comparing M2A to TDN, the largest improved classes were:
\begin{itemize}
  \item ``Poking something so that it falls over" (+21.9\%) 
  \item ``Moving away from something with your camera" (+20.4\%)
\end{itemize}

And lastly comparing M2A to TEA, the largest improved classes were:
\begin{itemize}
  \item ``Pretending to put something on a surface" (+16.7\%)
  \item ``Poking something so it slightly moves" (+12.5\%) 
\end{itemize}

We see that M2A improves on motion-oriented classes which include `moving' and `approaching' (i.e., ``Moving away from something with your camera") and interaction classes such as `poking something' (i.e., ``Poking something so it slightly moves"). This shows that incorporating motion information with attention mechanisms can improve the classification of motion-oriented classes and interaction-based classes in videos compared to using just motion or just attention mechanisms.

\begin{table}[t]
	\caption{Comparison of different state-of-the-art motion and attention mechanisms inserted into a ResNet18 backbone.}
	\label{table:state-of-the-art-ablation}
	\def\arraystretch{1.25}
	\centering
	\begin{tabular}{|l|l|l|l|}
		\hline
		\multirow{2}{*}{Motion Mechanism} & \multicolumn{3}{c|}{Top-1 Acc}                                      \\ \cline{2-4} 
		& None & M2A-Attention & TAM-Attention  \cite{Wu2021} \\ \hline
		None                    & 13.9  & 16.9   & 21.0          \\ \hline
		M2A-Motion                 & 28.9 & \textbf{34.7 (+17.8\%)}  & 31.6     (+10.6\%)     \\ \hline
		TDN-Motion     \cite{tdn}  & 28.6 & 29.0 (+12.1\%) & 25.3   (+4.3\%)        \\ \hline
		TEA-Motion    \cite{Li2020}    & \textbf{34.3} & 33.9 (+17.0\%) & \textbf{33.3  (+12.3\%)}      \\ \hline
	\end{tabular}
\end{table}

\subsection{Extending SOTA motion/attention-only mechanisms}

Lastly, we attempted to extend state-of-the-art motion-only and attention-only mechanisms by incorporating attention and motion respectively. The results are shown in Table \ref{table:state-of-the-art-ablation}. The first column states the motion mechanism used and the first row states the attention mechanism used. For example, the cell which intersects M2A-Motion and M2A-Attention is the full M2A mechanism. We also show the improvement made by incorporating motion with each attention mechanism in brackets and bold the highest accuracy for each attention mechanism. 

We see that all attention methods are improved when motion is incorporated (e.g, TAM-Attention achieves 21.0\% top-1 accuracy without motion and achieves 33.3\% top-1 accuracy when combined with TEA-motion). However, we also see that the motion mechanisms are not always improved when attention mechanisms are incorporated. Specifically, TDN and TEA mechanisms see a very small or negative change in accuracy when attention is added to them (e.g, TEA-Motion achieves 34.3\% top-1 accuracy without attention and 33.9\% with M2A-Attention). This may be because their mechanisms do something similar to attention mechanisms so incorporating attention mechanisms with them doesn't lead to any improvements.

\section{Conclusions}

In this paper, we introduce a new temporal mechanism, Motion Aware Attention (M2A), which utilizes both motion and attention for accurate video recognition. We showed that M2A can accurately recognize actions across multiple CNN backbones including 2D-ResNet18, 2D-MobileNet, and I3D-ResNet18 and that the proposed M2A mechanism can lead to a +15\% to +26\% improvement in top-1 accuracy with only a small increase in computational complexity. Furthermore, we showed how other SOTA attention mechanisms can be further improved by explicitly incorporating motion characteristics. Lastly, we also showed that M2A achieves competitive accuracy and efficiency compared to other SOTA temporal mechanisms and can lead to up to +60\% in top-1 accuracy across specific classes in SSV1. We hope this research helps develop more accurate and efficient temporal mechanisms for video action recognition. 

\bibliographystyle{abbrv}
\bibliography{m2a}

\end{document}